%% file: main.tex
\documentclass[journal]{IEEEtran}
\ifCLASSINFOpdf
\else
\fi

\usepackage{breqn}
\usepackage{makeidx}  
\usepackage{lipsum}
\usepackage{multirow}
\usepackage[usenames, dvipsnames]{color}
\usepackage{graphicx}
\usepackage{wrapfig}
\usepackage{url}
\usepackage{hhline}
\usepackage{caption}
\usepackage{cleveref}
\usepackage{soul}
\usepackage{hyperref}
\usepackage{subcaption}
\usepackage[para]{threeparttable}
\usepackage{booktabs}
\usepackage{siunitx}
\usepackage{verbatim}
\usepackage{tikz} 
\usepackage{tikz-qtree-compat}  
\usepackage[colorinlistoftodos]{todonotes}
\usepackage{rotating}

\begin{document}

%\title{Classification of food-related scenes in egocentric photostreams}
%\title{Bayesian semantic neural network for food scenes classification in egocentric images}

\title{Hierarchical approach to classify food scenes in egocentric photo-streams}

\author{Estefan\'ia Talavera, Mar\'ia Leyva-Vallina, Md. Mostafa Kamal Sarker, Domenec Puig, Nicolai Petkov and Petia Radeva %IAPR Fellow
%https://www.overleaf.com/1374734266myrfwydhbczk

%\author{Michael~Shell,~\IEEEmembership{Member,~IEEE,}
%        John~Doe,~\IEEEmembership{Fellow,~OSA,}
%        and~Jane~Doe,~\IEEEmembership{Life~Fellow,~IEEE}% <-this % stops a space
\IEEEcompsocitemizethanks{\IEEEcompsocthanksitem E. Talavera, M. Leyva-Vallina and N. Petkov are with the University of Groningen. \protect\\
E-mail: estefania.tama@gmail.com
\IEEEcompsocthanksitem  P. Radeva is with the University of Barcelona.
\IEEEcompsocthanksitem M.M.K. Sarker and D. Puig are with the University of Rovira i Virgili.}% <-this % stops a space

\thanks{Manuscript received x, 2018; revised x, 2018.}}

\markboth{IEEE XXX}
%\markboth{Journal of \LaTeX\ Class Files,~Vol.~xx, No.~x, October~2018}%
{Talavera \MakeLowercase{\textit{et al.}}: Hierarchical approach to classify food scenes in egocentric photo-streams}


\maketitle

\begin{abstract}

% \subsection{Background}
Recent studies have shown that the environment where people eat can affect their nutritional behaviour
\cite{howweeat}. In this work, we provide automatic tools for personalised analysis of a person's health habits by the examination of daily recorded egocentric photo-streams.
% \subsection{Aim}
Specifically, we propose a new automatic approach for the classification of food-related environments, that is able to classify up to 15 such scenes. In this way, people can monitor the context around their food intake in order to get an objective insight into their daily eating routine.
% \subsection{Method}
We propose a model that classifies food-related scenes organized in a semantic hierarchy. Additionally, we present and make available a new egocentric dataset composed of more than 33000 images recorded by a wearable camera, over which our proposed model has been tested.
% \subsection{Results}
Our approach obtains an accuracy and F-score of 56\% and 65\%, respectively, clearly outperforming the baseline methods.

\end{abstract}


\IEEEpeerreviewmaketitle

% SECTION %.....%.....%.....%.....%.....%.....%.....%.....%.....%.....%.....%.....%.....%.....%.....%.....

\section{Introduction}
\label{sec:introduction}

%\subsection{Background}
\IEEEPARstart{N}{utrition} is one of the main pillars of a healthy lifestyle. It is directly related to most chronic diseases like obesity, diabetes, cardiovascular diseases, and also cancer and mental diseases \cite{stalonas1985behavioral,hopkinson2006prevalence,donini2003eating}. Recent studies show that it is not only important \textit{what people eat}, but also \textit{how/where people eat} \cite{howweeat}. For instance, it is common knowledge that it is advised a person who is on a weight-reduction plan should to not go to the supermarket while being hungry \cite{Tal2013FatteningFood}. Social environment also matters; we eat more in certain situations, such as parties than at home \cite{eatparties}. If we are exposed to the food we feel the need or temptation to eat, the same feeling of temptation will be experienced at the supermarket \cite{Kemps2014ExposureEat}. Not only the sight plays its role, but also smell: everyone has walked in front of a bakery shop and felt tempted or hungry immediately \cite{FooDaroma}. The conclusion is that \textit{where we are} can have a direct impact on \textit{what or how we eat} and, by extension, on our health \cite{foodhealth}. However, there is a clear lack of automatic tools to monitor objectively the context of our food intake along time. 

% Aims: To understand how Recovery College students and tutors experience the design and delivery of a mental health Recovery College course, specifically the “‘Building Resilience” course.

\begin{figure}[ht!]
\centering
\includegraphics[width=0.47\textwidth]{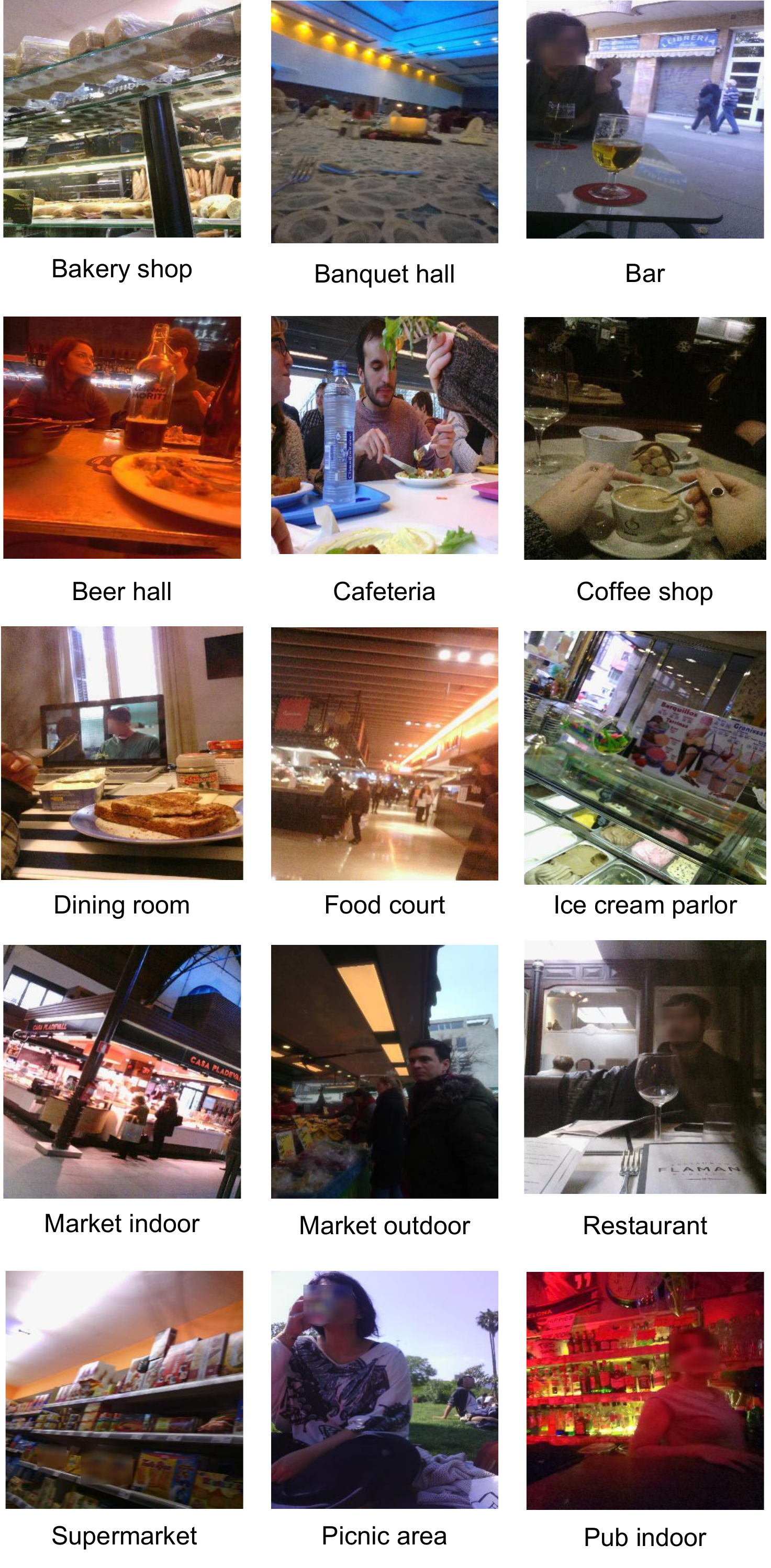}
\caption{Examples of images of each of the proposed food-related categories present in the introduced EgoFoodPlaces dataset.}
\label{fig:egoexamples}
\end{figure}

\subsection{Our aim}
Our aim is to propose an automatic tool based on robust deep learning techniques able to classify food-related scenes where a person spends time during the day. Our hypothesis is that if we can help people get insight into their daily eating routine, they can improve their habits and adopt a healthier lifestyle. By \textit{eating routine}, we refer to activities related to the acquisition, preparing and intake of food, that are commonly followed by a person. For instance, `after work, I go shopping and later I cook dinner and eat'. Or, `I go after work directly to a restaurant to have dinner'. These two eating routines would affect us differently, having a direct impact on our health. The automatic classification of food-related scenes can represent a valuable tool for nutritionists and psychologists as well to monitor and understand better the behaviour of their patients or clients. This tool would allow them to infer how the detected eating routines affect the life of people and to develop personalized strategies for behaviour change related to food intake. 

% \subsection{Research context}
The closest approaches in computer vision to our aim focus either on scene classification, with a wide range of generic categories, or on food recognition from food-specific images, where the food typically occupies a significant part of the image. However, food recognition from these pictures does not capture the context of food intake and thus does not represent a full picture of the routine of the person. It mainly exposes what the person is eating, at a certain moment, but not \textit{where, in which environment}. These environmental aspects are important to analyze in order to keep track of the people behaviour. 

\subsection{Personalized Food-Related Environment Recognition}
In this work, we propose a new tool for the automatic analysis of food-related environments of a person. In order to be able to capture these environments along time, we propose to use recorded egocentric photo-streams. These images provide visual information from a first-person perspective of the daily life of the camera wearer by taking pictures frequently: visual data about activities, events attended, environments visited, and social interactions of the user are stored. Additionally, we present a new labelled dataset that is composed of more than 33000 images, which were recorded in 15 different food-related environments. 

The differentiation of food-related scenes that commonly appear in recorded egocentric photo-streams is a challenging task due to the need to recognize places that are semantically related. In particular, images from two different categories can look very similar, although being semantically different. Thus, there exists a high inter-class similarity, in addition to a low intra-class variance (i.e. semantically similar categories, like \textit{restaurant} and \textit{pizzeria}, might look visually similar). In order to face this problem, we consider a taxonomy taking into account the relation of the studied classes. The proposed model for food-related scene classification is a hierarchical classifier that embeds convolutional neural networks emulating the defined taxonomy. 

The contributions of the paper are three-fold:
\begin{itemize}

\item A deep hierarchical network for classification of food-related scenes from egocentric images. The advantage of the proposed network is that it adapts to a given taxonomy. This allows the classification of a given image into several classes describing different levels of abstraction.

\item  A taxonomy of food-related environments organized in a fine-grained way that takes into account the main food-related activities (eating, cooking, buying, etc.). Our classifier is able to classify the different categories and subcategories of the taxonomy within the same model.
\item An egocentric dataset of 33000 images and 15 food-related environments. We call it FoodEgoPlaces and, together with its ground-truth, is publicly available in \url{http://www.ub.edu/cvub/dataset/.}
\end{itemize}

%\textcolor{blue}{As an example of application, we illustrate the utility of the proposed method for the detection of food-related scenes.} 

% Paper organization
% \subsection{Paper organization}
The paper is organized as follows: in Section \ref{section:PreviousWorks}, we highlight some relevant works related to our topic, in Section \ref{section:proposedmodel} we describe the approach proposed for food scene recognition. In Section \ref{section:experimentalsetup}, we introduce our FoodEgoPlaces dataset and outline the experiments performed and obtained results. In Section \ref{section:Discussions}, we discuss the results achieved. Finally, in Section \ref{section:conclusions}, we present our conclusions. 

% SECTION %.....%.....%.....%.....%.....%.....%.....%.....%.....%.....%.....%.....%.....%.....%.....%.....

\section{Previous Works}
\label{section:PreviousWorks}

Scene recognition has been extensively explored in different fields, namely: robotics\cite{falomir2012qualitative}, surveillance \cite{makris2005learning} environmental monitoring\cite{higuchi2011imaging} or egocentric videos \cite{cartas2018batch}. In this section, we describe previous works addressing this topic. 

The recognition and monitoring of food-intake have been previously addressed in the literature \cite{fontana2014automatic,ravi2015real,liu2012intelligent}. For instance, in \cite{fontana2014automatic}, the authors proposed the use of a microphone and a camera worn on the ear to get insight into the subject's food intake. On one side, the sound allows the classification of chewing activities, and on the other side, the selection of keyframes create an overview of the food intake that otherwise would be difficult to quantify. A food-intake log supported by visual information allows inferring the food-related environment where a person spends time. However, no work has focused on this challenge so far.

\subsection{Scene classification}
The problem of scene classification was originally addressed in the literature by applying traditional techniques (\cite{Lazebnik2006BeyondCategories,Quattoni2009RecognizingScenes.}, just to mention a few), over handcrafted features. Nowadays, deep learning is the state-of-the-art \cite{zhou2017places}.

As for the former case, one of the latest works on scene recognition using traditional techniques is  \cite{Lazebnik2006BeyondCategories}, whose aim was to recognize 15 different scenes categories of outdoor and indoor scenes. The proposed model was based on the analysis of image sub-region geometric correspondences by computing histograms of local features. In \cite{Quattoni2009RecognizingScenes.}, the proposed approach focused on indoor scenes recognition, extending the number of recognized scenes to 67, where 10 of them are food-related. Having the hypothesis that similar scenes contain specific objects, their approach combines local and global image features for the definition of prototypes for the studied scenes. Very soon scene recognition was outperformed using deep learning.

% \subsection{Convolutional Neural Networks and big datasets of scene images}
Convolutional Neural Networks (CNNs) are a type of feed-forward artificial neural network with specific connectivity patterns. %were inspired by the animal's visual cortex neurons connections \cite{rosenblatt1958perceptron}. 
Since Yann LeCun's LeNet \cite{Lecun1998Gradient-BasedRecognition} was introduced, many other deep architectures have been developed and applied to different computer vision known problems, achieving better results than the state-of-art techniques: MNIST \cite{Lecun1998Gradient-BasedRecognition} (images), Reuters \cite{LewisReuters-21578}(documents) and TIMIT \cite{Garofolo1993TIMITCorpus} (recordings in English), ImageNET \cite{DengImageNetDatabase} (Data Sets classification), etc.  Within the wide range of recently proposed architectures, some of the most popular are: GoogleNet \cite{Szegedy2015GoingConvolutions}, AlexNet \cite{Krizhevsky2012ImageNetNetworks}, ResNet \cite{He2016DeepRL}, or VGGNet \cite{Simonyan2015VeryRecognition}.  
The use of CNNs for learning high-level features has shown huge progress in scene recognition outperforming traditional techniques like \cite{Quattoni2009RecognizingScenes.}. This is mostly due to the availability of large datasets, those presented in \cite{Quattoni2009RecognizingScenes.,Yu_lsun:construction} or the ones derived from the MIT Indoor dataset (\cite{Zhou2014LearningDatabase,Zhou2016Places:Understanding}). However, the performance at \textit{scene recognition} level has not reached the same level of success as \textit{object recognition}. Probably, this is a result of the difficulty presented when generalizing the classification problem, due to the huge range of different environments surrounding us (e.g. 400 in the Places2 dataset \cite{Zhou2014LearningDatabase}). 

In \cite{KoskelaConvolutionalRecognition}, CNN activation features were extracted and concatenated following a spatial pyramid structure and used to train one-vs-all linear classifiers for each scene category. In contrast, in \cite{Zhou2014LearningDatabase} the authors evaluate the performance of the responses from the trained Places-CNN as generic features, over several scene and object benchmarks. Also, a probabilistic deep embedding framework, which analyses regional and global features extracted by a neural network, is proposed in \cite{Zheng2014SeeingEvidences}. In \cite{Wang2015Object-SceneImages}, two different networks called Object-Scene CNNs,  are combined by late fusion; the `object net' aggregates information for event recognition from the perspective of objects, and the `scene net' performs the recognition with help from the scene context. The nets are pre-trained on the ImageNet dataset \cite{DengImageNetDatabase} and Places dataset \cite{Zhou2014LearningDatabase} respectively. Recently, in \cite{Herranz2016SceneBias} the authors combine object-centric and scene-centric architectures. They propose a parallel model where the network operates over different scale patches extracted from the input image. None of these methods has been tested on egocentric images, which by themselves represent a challenge for image analysis. In this kind of data, the camera follows the user's movements. This results in big variability on illumination, blurriness, occlusions, drastic visual changes due to the low frame rate of the camera, narrow field of view, among other difficulties.  
%neither just on food-related scene classification. 
% Hence, we consider that there is not direct comparison to them.

\subsection{Classification of egocentric scenes}
 In order to obtain personalized scene classification, we need to analyze egocentric images acquired by a wearable camera. Egocentric image analysis is a relatively recent field within computer vision concerning the design and development of Computer Vision algorithms to analyze and understand photo-streams captured by a wearable camera. In \cite{B2016TemporalInterest}, several classifiers were proposed to recognize 8 different scenes (not all of them food-related). First, they discriminate between food/no-food and later, they train One-vs-all classifiers to discriminate among classes. Later, in \cite{Furnari2017RecognizingVideos} a multi-class classifier was proposed, with a negative-rejection method applied. In \cite{B2016TemporalInterest,Furnari2017RecognizingVideos} they only consider 8 scene categories, just 2 of them are food-related (\textit{kitchen} and \textit{coffee machine}) and without visual or semantic relation. 
 
\subsection{Food-related scene recognition in egocentric photo-streams}
In our preliminary work presented in \cite{sarker2018macnet}, we proposed a MACNet neural architecture for the classification of food-related scenes. This network input image is scaled into five different resolutions (the original image, with a scale value of 0.5). The five scaled images are fed to five blocks of atrous convolutional networks~\cite{chen2018deeplab} with three different rates (1, 2, and 3) to extract the key features of the input image in multi-scale. In addition, four blocks of pre-trained ResNet are used to extract 256, 512, 1024 and 2048 feature maps, respectively. Each feature maps extracted by an atrous convolutional block is concatenated with the corresponding ResNet block to feed the subsequent block. Finally, the features obtained from the fourth ResNet layer is the final features are used to classify the food places images using two fully connected (FC) layers.

% into 22 categories. 
However, the challenge still remains due to the high variance that environments take in real-world places and the wide range of possibilities of how a scene can be captured. In this work, we propose an organization of the different studied classes into semantic groups following the logic that relates them. We define a taxonomy, i.e. a semantic hierarchy relating the food-related classes. Hierarchical classification is an iterative process that groups features or concepts based on their similarity into clusters, until merging them all together. There are two strategies for hierarchical classification: agglomerative (bottom-up) and divisive (top-down). We aim to classify food-related images following a top-down strategy, i.e. from a less to a more specific description of the scene. The proposed hierarchical model supports its final classification on the dependence among classes at the different levels of the classification tree. This allows us to study different levels of semantic abstraction. The different semantic levels (L), Level 1 (L1), Level 2 (L2) and Level 3 (L3),  are introduced in Fig. \ref{fig:baynet} In this document we refer to meta-class as the class whose instances are semantic and visual correlated classes.

Therefore, we organize environments according to the actions related to them: cooking, eating, acquiring food products. We demonstrate that by creating different levels of classification and classifying scenes by the person action, it can serve as a natural prior for more specific environments and thus can further improve the performance of the model. The proposed classification model, implemented following this taxonomy, allows analyzing at different semantic levels of where the camera wearer spends time.

To the best of our knowledge, no previous work has focused on the problem of food-related scenes recognition at different semantic levels, either from conventional or egocentric images. Our work aims to classify food-related scenes from egocentric images recorded by a wearable camera. We believe that these images highly describe our daily routine and can contribute to the improvement of healthy habits of people.

\section{Hierarchical approach for food-related scenes recognition in egocentric photo-streams}
%Approach}
\label{section:proposedmodel}

We propose a new model to address the classification of food-related scenes in egocentric images. It follows a hierarchical semantic structure, which adapts to the taxonomy that describes the relationships among classes. The classes are hierarchically implemented from more abstract to more specific ones. Therefore, the model is scalable and can be adapted depending on the classification problem, i.e. if the taxonomy changes. 

\begin{figure}[ht!]
\centering
\includegraphics[width=0.47\textwidth]{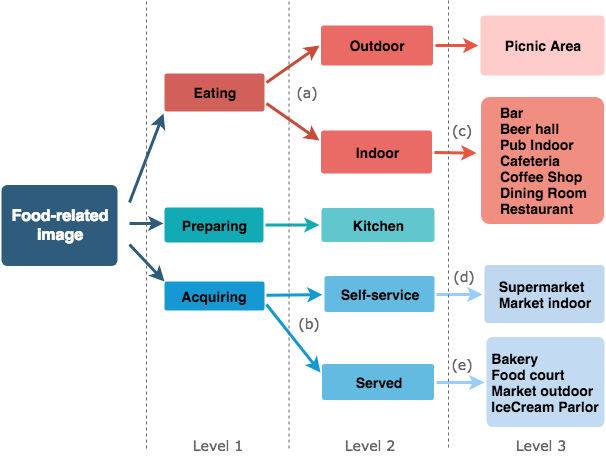}
\caption{The proposed semantic tree for food-related scenes categorization. For their later reference, we mark with dashed lines the different depth levels, and with letters the sub-classification groups.}
\label{fig:baynet}
\end{figure}

For the purposes of food-related scene classification, we define a semantic tree which is depicted in Fig. \ref{fig:baynet}. We redefine the problem inspired by how humans hierarchically organize concepts into semantic groups. The Level 1 directly related to the problem of physical activities recognition \cite{cartas2018batch}: \textit{eating}, \textit{preparing}, and \textit{acquiring food (shopping)}. Note that the recognition of physical activities itself is a well-known and still open research problem in egocentric vision\cite{cartas2018batch}. On the other hand, recognition of these three activities has multiple applications like for patients with Mild Cognitive Impairment (MCI) in the Cambridge cognition test \cite{schmand2000early}.  There, the decrease of older people’s cognitive functions with time is one of the factors to estimate their cognitive capacities by measuring their capacity to prepare food or go for shopping \cite{petersen1999mild}. Later it splits eating into eating outdoor or indoor. Some of the subcategories group several classes, such as the subcategory \textit{eating indoor} that encapsulates seven food-related scenes classes: \textit{bar, beer hall, cafeteria, coffee shop, dining room, restaurant,} and \textit{pub indoor}. In contrast, \textit{preparing} and \textit{eating outdoor} are represented uniquely by \textit{kitchen} and \textit{picnic area}, respectively. The semantic hierarchy was defined following the collected food-related classes and their intrinsic relation. Thus, the automatic analysis of the frequency and duration of such food-related activities is of high importance when analyzing their behaviour. The environment is differentiated in Level 2. As commented in the manuscript, in \cite{howweeat} the authors stated that ‘where you are, affects your eating habits’. Thus, the food routine or habits of camera wearers can be inferred by recognizing the food-related environment where they spend time (e.g. outdoor, indoor, etc.).

The classification of scenes is already a scientific challenge, see the dataset Places \cite{zhou2017places}. For us, the novelty is to address the classification of scenes with similar characteristics (food-related) that makes the problem additionally more difficult. 

We proposed this taxonomy because we think it represents a powerful tool to address the behaviour of people. Moreover, it could be of interest in order to estimate the cognitive state of MCI patients. We reached this conclusion after previous collaborations with psychologists working on the MCI disorder, and analysing egocentric photo-streams addressing several problems.

The differentiation among classes at the different levels of the hierarchy needs to be performed by a classifier. In this work, we propose to use CNNs for the different levels of classification of our food-related scenes hierarchy. The aggregation of CNNs layers mimics the structure of the food-related scenes presented in Fig.  \ref{fig:baynet}. Due to the good quality of the scene classification results over the Places2 dataset \cite{Zhou2016Places:Understanding}, we made use of the pre-trained VGG16 introduced in \cite{Simonyan2015VeryRecognition}, on which we built our hierarchical model. In this work, we will refer to it as VGG365 network. Note that this approach resembles the DECOC classifier \cite{DBLP:journals/pami/PujolRV06} that proves the efficiency of decomposing a multi-class classification problem in several binary classification problems organized in a hierarchical way. The difference with the food-related scene classification is that in the latter case the classes are organized semantically in meta-classes corresponding to nutrition-related activities instead of constructing meta-classes without explicit meaning, but according to the entropy of training data \cite{DBLP:journals/pami/PujolRV06}.

Given an image, the final classification label is based on the aggregation of estimated intermediate probabilities obtained for the different levels of the hierarchical model, since a direct dependency exists between levels of the classification tree. The model aggregates the chain of probabilities by following the statistical inference method. The probability of an event is based on its prior estimated probabilities.

Let us consider classes $C^i$ and $C^{i-1}$ so that superscript shows the level of the class in the hierarchy and $C^{i-1}$ is the parent of $C^i$ in the hierarchical organization of the tree. Thus, we can write:
\begin{equation}
\begin{normalsize}
P(C^i,x) = P(C^i,x|C^{i-1},x)*P(C^{i-1}|x)
\end{normalsize}
\end{equation}
where $P()$ relates to probabilities. $P(C^{i-1},x|C^i,x)$ represents the likelihood of $C^{i-1}$, given image x, occurring given that $C^i$, given image x, is happening, while $P(C^i,x)$ and $P(C^{i-1},x)$ are marginal probabilities given image x, i.e. the probabilities of independently observing $C^i$ and $C^{i-1}$, respectively.

Note that we can estimate $P(C^i,x|C^{i-1},x)$ from the classifier of the network trained to classify the classes children of class $C^i$, $P(C^{i-1},x|C^i,x)$ is 1 since $C^i$ is a subclass of $C^{i-1}$. 

$P(C^{i-1},x)$ can be recursively estimated by considering the estimated probability on $C^{i-1}$ and its class parent. Hence, we obtain that for each node $C^i$ in the hierarchy (in particular, for the leaves), we get:

\begin{equation}
\begin{normalsize}
P(C^i,x) = \Pi_{j=1}^i P(C^j,x|C^{j-1},x)*P(C^{j-1},x)
\end{normalsize}
\end{equation}

Without loss of generality, we consider that the probability of the class in the root is the probability to have a food-related image, ($P(C^0)$), obtained by a binary classifier. 

Let us illustrate the process with an example. Following the semantic tree in Fig. \ref{fig:baynet}, our goal is to classify an egocentric image belonging to the class \textit{dining room}. We observe that as \textit{dining room} is a subclass of  \textit{indoor} and \textit{indoor} is of \textit{eating}, etc. Thus, the probability of \textit{dining room} occurring giving image \textit{x} is computed as:

\begin{dmath}
P(diningroom,x)=  P(dining room,x|indoor,x) * P(indoor,x|eating,x)*P(eating,x|food related,x)*P(food related,x)
\end{dmath}

To summarize, given an image, our proposed model computes the final classification as a product of the estimated intermediate probabilities at the different levels of the defined semantic tree. 

%%
%% ... 
%% ----------------------------------------
%% ... 
%%
% SECTION %.....%.....%.....%.....%.....%.....%.....%.....%.....%.....%.....%.....%.....%.....%.....%.....
%.....%.....%.....%.....%.....%.....%.....%.....%.....%.....%.....%.....%.....%.....%.....%.....%.....%.....

\section{Experiments and Results}
\label{section:experimentalsetup}

In this section, we describe a new home-made dataset that we make public, the experimental setup, the metrics used to evaluate the analysis, and the obtained results.

\subsection{Dataset}

In this work, we present \textit{EgoFoodPlaces}, a dataset composed of more than 33000 egocentric images from 11 users organized in 15 food-related scene classes. The images were recorded by a Narrative Clip camera\footnote{http://getnarrative.com/}. This device is able to generate a huge number of images due to its continuous image collection. It has a configurable frame rate of 2-3 images per minute. Thus, users regularly record an amount of approximately 1500 images per day. The camera movements and the wide range of different situations that the user experiences during his/her day, lead to new challenges such as background scene variation, changes in lighting conditions, and handled objects appearing and disappearing throughout the photo sequence.

Food-related scene images tend to have an intrinsic high inter-class similarity, see Fig. \ref{fig:egoexamples}. To determine the food-related categories, we selected a subset of the ones proposed for the Places365 challenge \cite{zhou2017places}. We focus on the categories with a higher number of samples in our collected egocentric dataset, disregarding very unlikely food-related scenes, such as \textit{beer garden} and \textit{ice-cream parlor}. Furthermore, we found that discriminating scenes like \textit{pizzeria} and \textit{fast-food restaurant} is very subjective if the scene is recorded from a first-person view, and hence, we merged them into a \textit{restaurant} class. 

\begin{figure}[ht!]
\centering
\includegraphics[width=0.5\textwidth]{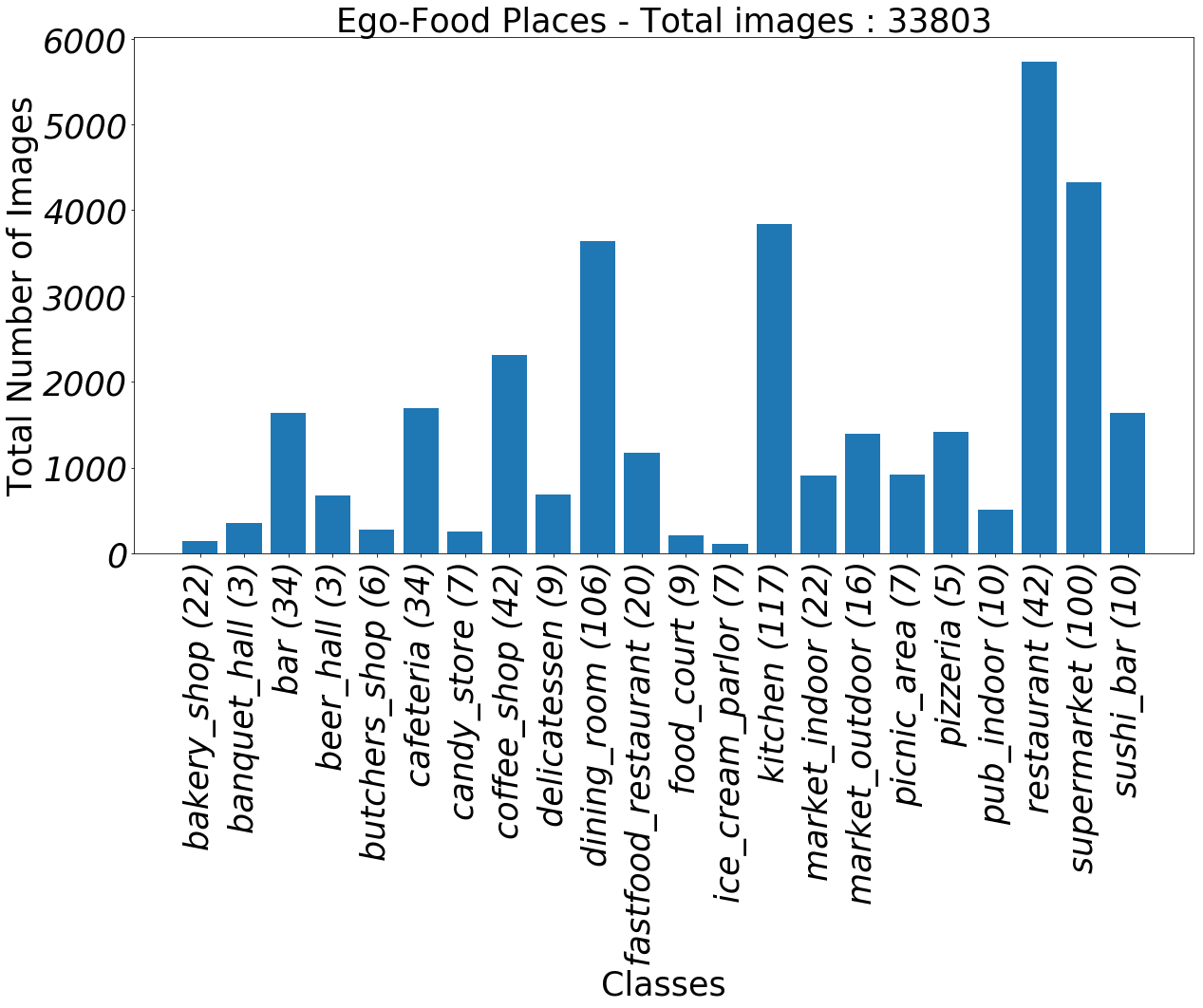}
\caption{Total number of images per food-related scene class. We give the number of collected events per class between parenthesis.}
\label{fig:datasetNumbers}
\end{figure}

\textit{EgoFoodPlaces} was collected during the daily activities of the users. To build the dataset, we select the subset of images from the EDUB-Seg dataset that described food-related scenes, introduced in \cite{Talavera2015R-clusteringSegmentation,Dimiccoli2015SR-Clustering:Segmentation}, and later extended it with new collected frames. The dataset was gathered by 11 different subjects, during a total of 107 days, while spending time in scenes related to the \textit{acquisition, preparing} or \textit{consumption} of food.  The dataset has been manually labelled into a total of 15 different food-related scenes classes: \textit{bakery, bar, beer hall, cafeteria, coffee shop, dining room, food court, ice cream parlour, kitchen, market indoor, market outdoor, picnic area, pub indoor, restaurant, and supermarket.} In Fig. \ref{fig:datasetNumbers}, we show the number of images per different classes. This figure shows the unbalanced nature of the classes in our dataset, reflecting the different prolongation of time that a person spends on different food-related scenes.

\begin{figure}[ht!]
\centering
\includegraphics[width=0.5\textwidth]{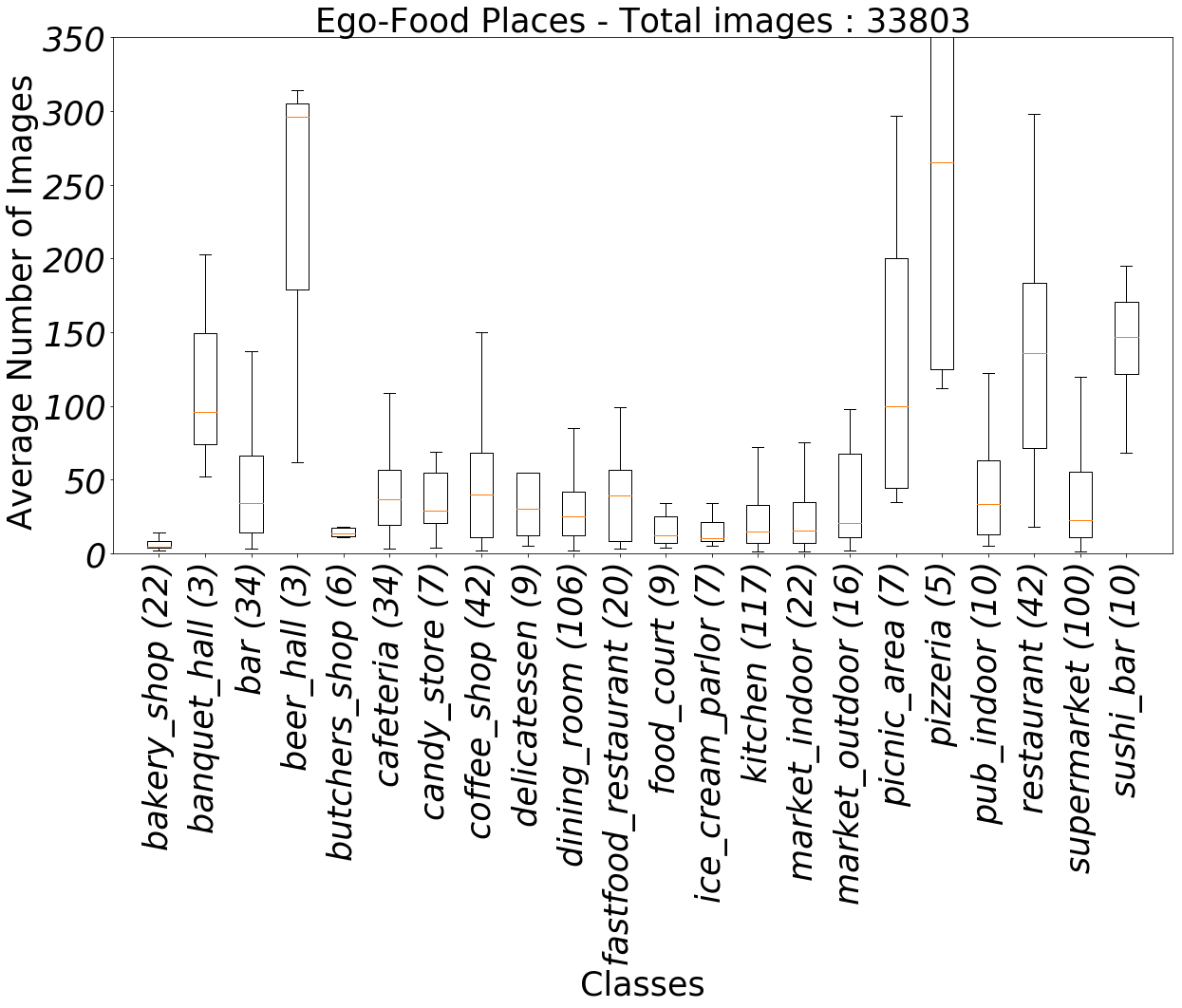}
\caption{Illustration of the variability of the size of the events for the different food-related scene classes. The data is presented by making the width of the box proportional to the size of the group. We give the number of collected events per class between parenthesis. The range of the data of a class is shown by the whiskers extend from its data box.}
\label{fig:variability}
\end{figure}

Since the images were collected by a wearable camera when performing any of the above-mentioned activities, the dataset is composed of groups of images close in time. This leads to two possible situations. On one hand, images recorded `sitting in front of a table while having dinner' will most likely be similar. On the contrary, in scenes such as `walking at the supermarket' the images vary since they follow the walking movement of the user in a very varying environment. 

\begin{comment}
\begin{figure*}[ht!]
\centering
\input{figure_t-sne.tex}
\caption{Visualization of the distribution of the classes using the t-SNE algorithm.}
\label{fig:t-SNE}
\end{figure*}
\end{comment}

\begin{figure}[ht!]
\centering
\input{figure_t-sne.tex}
\caption{Visualization of the distribution of the classes using the t-SNE algorithm.}
\label{fig:t-SNE}
\end{figure}
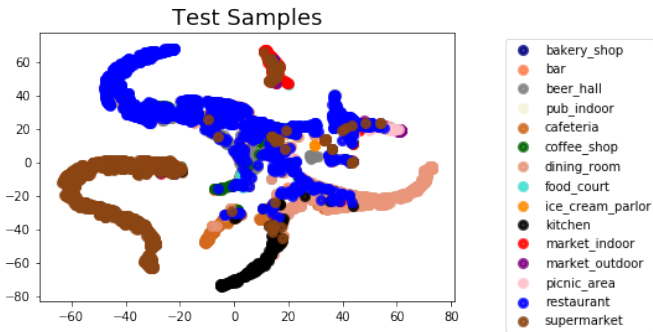

In Fig. \ref{fig:variability}, we present the dataset by classes and events. This graph shows how the average, maximum and minimum spent time for the given classes differ. Note that this time can be studied since it is directly related to the number of recorded images in the different food-related scenes. As we previously assumed, classes with a small number of images correspond to unusual environments or environments where people do not spend a lot of time in (e.g. \textit{bakery}). In contrast, the most populated classes refer to everyday environments (e.g. \textit{kitchen, supermarket}), or to environments where more time is usually spent (e.g. \textit{restaurant}). 

\subsubsection{Class-variability of the EgoFoodPlaces dataset}

%\begin{figure}[ht!]
%\includegraphics[width=0.5\textwidth]{Figures/silh_score_.png}
%\label{fig:Silhouette_core}
%\end{figure}

\begin{figure}
\centering
\begin{subfigure}[b]{0.5\textwidth}
   \includegraphics[width=1\linewidth]{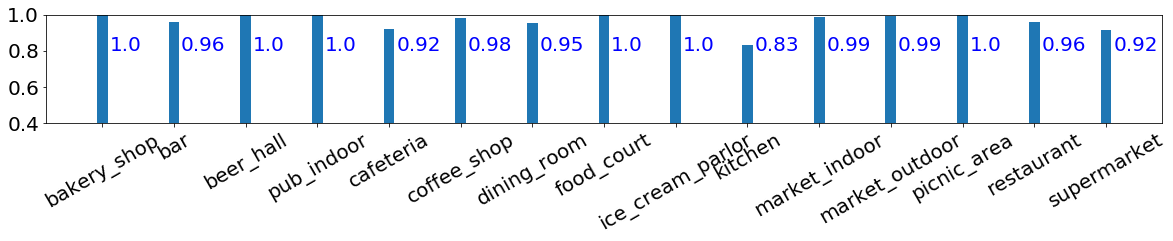}
   \caption{}
   \label{fig:Ng1} 
\end{subfigure}
\begin{subfigure}[b]{0.5\textwidth}
   \includegraphics[width=1\linewidth]{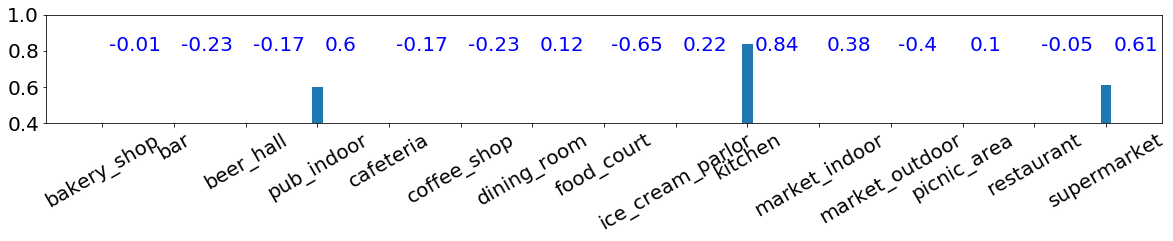}
   \caption{}
   \label{fig:Ng2}
\end{subfigure}
\caption{Mean Silhouette Score for the samples within the studied food-related classes. The train and test sets are evaluated separately in (a) and (b), respectively. The score is shown with bars and in blue text on top of them.}
\label{fig:Silhouette_core}
\end{figure}

To quantify the degree of semantic similarity among the classes in our proposed dataset, we compute the intra- and inter-class correlation. We use the classification probabilities output of the proposed baseline VGG365 network in order to find suitable descriptors for our images for this comparison. This network was trained for the classification of the proposed 15 food-related scenes. These descriptors encapsulate the semantic similarities of the studied classes.

To study the intra-class variability, we compute the mean silhouette coefficient for all samples, that is defined as,
\begin{equation}
Silhouette\_score = (b - a) / max(a, b)
\end{equation}
where $(a)$ corresponds to the intra-class distance per sample, and $(b)$ corresponds to the distance between a sample and the closest class to which the sample is part of. Note that the silhouette takes values from 1 to -1; the highest value represents high density and separated clusters. The value 0 represents overlapping of clusters. Negative values indicate that there are samples with more similar clusters than the one they have been assigned to. The mean Silhouette score is 0.94 and 0.15 for the train and test samples, respectively. The score is depicted for the different analyzed classes in Fig. \ref{fig:Silhouette_core}. The high score obtained for the train set is due to the fact that the analyzed descriptors are extracted fine-tuning the network with those specific samples. Thus, their descriptors are of high quality for their differentiation. In contrast, the test set is an unseen set of images. The low value of the test set indicates that the classes are challenging to classify.

\begin{comment}
\begin{table*}[t!]
\begin{center}
\caption{Split of the proposed EgoFoodPlaces dataset for the performed experiments.}
\vspace{1cm}
\label{table:splitDataset}
\input{splitDataset}
\end{center}
\end{table*}
\end{comment}
Furthermore, we visually illustrate the inter-class variability of the classes by embedding the 15-dimensional descriptor vector to 2 dimensions using the t-SNE algorithm \cite{maaten2008visualizing}. The results are shown in Fig. \ref{fig:t-SNE}.  This visualization allows us to better explore the variability among the samples in the test set. For instance, classes such as \textit{restaurant} and \textit{supermarket} are clearly distinguishable as a cluster. In contrast, we can recognize the classes with lower recognition rate, like the ones overlapping with \textit{supermarket} and \textit{restaurant}. For instance, \textit{market indoor} is merged in its majority with \textit{supermarket}. At the same time, the class \textit{restaurant} clearly overlaps with \textit{coffee shop} and \textit{picnic area}.

%%
%% ... ----------------------------------------
%% --------------------------------------------------------------------------------
%% ... ----------------------------------------
%%
\subsection{Experimental setup} 
\label{Subsection:experiments}

In this work, we propose to build the model on top of the VGG365 network \cite{zhou2017places} since it outperformed state-of-the-art CNNs when classifying conventional images into scenes. We selected this network because it was already pre-trained with images describing scenes, and after evaluating and comparing its performance to the state-of-the-art CNNs. The classification accuracy obtained by the VGG16\cite{Simonyan2015VeryRecognition}, InceptionV3\cite{DBLP:journals/corr/SzegedyVISW15}, and ResNet50\cite{He2015_resnet50}, were 55.07\%, 51.22\%, and 60.43\%, respectively, lower than the 64.02\% accuracy achieved by the VGG365 network.

We build our hierarchical classification model by aggregating VGG365 nets over different subgroups of images/classes, emulating the proposed taxonomy for food-related scenes recognition in Fig.$\;$\ref{fig:baynet}. The final probability of a class is computed by the model, as described in Section \ref{section:proposedmodel}. 

The model adapts to an explicit semantic hierarchy that aims to classify a given sample of food-related scenes. Moreover, it aims to further understanding of the relation among the different given classes. Therefore, we compare the performance of the proposed model against existent methodologies that can be adapted to obtain similar classification information.

We compare the performance of the proposed model with the following baseline experiments:

\begin{enumerate}

    \item FV: Fine-tuning of the VGG365 network with \textit{EgoFoodPlaces}.

    \item FV-RF: We use this categorical distribution obtained by the fine-tuned VGG365 in (1) as image descriptors. Later, we train the Random Forest classifier with 200 trees \cite{Ho:1995:RDF:844379.844681}.
    
    \item FV-SVM: Fine-tuned VGG365 to obtain image descriptors and Support Vector Machines \cite{cortes1995support}. 
    
    \item FV-KNN: Fine-tuned VGG365 to obtain image descriptors and k-Nearest Neighbors \cite{altman1992introduction} (n=3). 
    
    \item SVM-tree: We use the categorical distribution obtained by the fine-tuned VGG365 as images descriptors of the subsets of images that represents the nodes of the tree. Later, we train SVM as nodes of the proposed taxonomy.
    
    \item MACNet \cite{sarker2018macnet}: We fine-tuned the MACNet network introduced in \cite{sarker2018macnet} to fit our proposed dataset.
 
    \item FV-Ensemble: We evaluate the performance of a stack of FV networks that are trained with a different random initialization of the final fully connected weights for classification. The final prediction is the average of the predictions of the networks. We ensemble the same number of CNNs as the number of CNNs included in the proposed hierarchical model, i.e. 6 CNNs.
    
\end{enumerate}

We perform a 3-Fold cross-validation of the proposed model to verify its ability to generalize and report the average value. The baseline methodologies are also evaluated following a 3-Fold cross-validation strategy.

We make use of the Scikit-learn machine learning library available for Python for the training of the traditional classifiers (SVM. RF, and KNN). For all the experiments, the images are re-sized at size 256x256. For the CNNs, we fine-tuned the baseline CNNs for 10 epochs, with a training batch size of 8, and run the validation set each 1000 iterations. The training of the CNNs was implemented using Caffe \cite{Jia2014Caffe:Embedding} and its Python interface. The code for the implementation of our proposed model is publicly available in \url{https://github.com/estefaniatalavera/Foodscenes_hierarchicalmodel}.

%%
%% ... 
%% ----------------------------------------
%% ... 
%%
\subsection{Dataset Split} In order to robustly generalize the proposed model and fairly test it, we assure that there are no images from the same scenes/events in both training and test sets. To this aim, we divide the dataset into events for the training and evaluation phases. Events are captured by sequentially recorded images that describe the same environment, and we obtain them by applying the SR-Clustering temporal segmentation method introduced in \cite{Dimiccoli2015SR-Clustering:Segmentation}. The division of the dataset into training, validation and test, aims to maintain a 70\%, 10\% and 20\% distribution, respectively. As it can be observed in Fig. \ref{fig:datasetNumbers}, \textit{EgoFoodPlaces} presents highly unbalanced classes. In order to face this problem, we could either subsample classes with high representation, or add new samples to the ones with low representation. We decided not to discard any image due to the relatively small number of images within the dataset. Thus, we balanced the classes for the training phase by over-sampling the classes with fewer elements. The training process of the network learns from randomly crops of the given images, the over-sampling simply passes the same instances several times, until reaching the defined number of samples per class, which will correspond to the number of samples of the most frequent class. For all the experiments performed, the images used for the training phase are shuffled in order to give robustness to the network. Together with the EgoFoodPlaces dataset, the given labels, and the training, validation and test files are publicly available for further experimentation (\url{http://www.ub.edu/cvub/dataset/}). 

%%
%% ... 
%% ----------------------------------------
%% ... 
%%
\subsection{Evaluation}
\label{metrics:averageaccuracy}

We evaluate the performance of the proposed method and compare it with the baseline models by computing the accuracy, precision, recall and $F_1$ (F-score). We calculate them per each class, together with their 'macro' and 'weighted' mean. 'Macro' calculates metrics for each label, and find their unweighted mean, while 'weighted' takes into account the true instances for each label. We also compute the weighted accuracy. The use of weighted metrics aims to face the unbalanced of the dataset, and intuitively expresses the strength of our classifier. This metric normalizes based on the number of samples per class.

The \textit{$F_1$ score}, \textit{Precision} and \textit{Recall} can be defined as:

\begin{equation}
F_1 = 2 \times  \frac{Precision * Recall}{ Precision + Recall},
\end{equation}
\begin{equation}
Precision = \frac{TruePositive}{ TruePositive+ FalsePositive},
\end{equation}
\begin{equation}
Recall = \frac{TruePositive}{TruePositive+ FalseNegative}.
\end{equation}

%Thus, one missed label will weigh more if we have less instances of the evaluated class, and vice-versa:
%\[
%w_i = \frac{1}{n_i*C}.
%\]
%For a class $i$ with $n_i$ instances, a hit or a miss will weigh $w_i$, if we have $C$ different classes.

Moreover, we qualitatively compare the given labels by our method and the best of the proposed baseline to sample images from the test set. 

%%
%% ... 
%% ----------------------------------------
%% ... 
%%
\subsection{Results}

We present the obtained classification accuracy at image level for the performed experiments in Table \ref{table:activityClassificationPerformance}. As it can be observed, our proposed model achieves the highest accuracy and weighted average accuracy, with 75.46\% and 63.20\%, respectively, followed by the SVM and Random Forest for the accuracy and SVM and KNN for the weighted accuracy. 

Our proposed hierarchical model has the capability of recognizing not only the 15 classes corresponding to the leaves of the tree in the semantic tree (see Fig.\ref{fig:baynet}), but also the meta-classes at the different semantic levels of depth. Thus, specialists can analyze the personal data and generate strategies for the improvement of the lifestyle of people by studying their food-related behaviour either from a broad perspective, such as when the person \textit{eats} or \textit{shops}, or into a more detailed one, like \textit{if the person usually eats in a fast-food restaurant or at home}.

A logical question is if the model provides a robust classification of meta-classes as well. To this aim, we evaluate the classification performance at the different levels of the defined semantic tree. Note that since each class is related to a meta-class on a higher level, an alternative to our model would be to obtain the meta-classes accuracy from their sub-classes classification. 
We compare the accuracy of meta-classes from their classification by the proposed model vs inferring the accuracy from the classification of the subclasses samples for the set of baseline models. As one can observe in Table \ref{table:tableLayersTree}, our model achieves higher accuracy classifying meta-classes in all cases with 94.7\%, 68.5\%, 94.7\% for Level 1 (L1), Level 2 (L2) and Level 3 (L3), respectively. This proves that it is a robust tool for the classification of food-related scenes classes and meta-classes.

\begin{table*}[ht!]
\begin{center}
\caption{Food-related scene classification performance. We present the accuracy per class and model, and precision, recall and F1 score for all models. We rename the fine-tuning of the VGG365 as `FV', and the later use of its output probabilities for the training of the State-of-the-Art models.}
\label{table:activityClassificationPerformance}
%\resizebox{1.0\columnwidth}{!}{%
\input{table_acc_results}

%}
\end{center}
\end{table*}

\begin{comment}

\begin{table*}[ht!]
\begin{center}
\caption{Average results for the 3 folds.}
\label{table:averageresults}
%\resizebox{1.0\columnwidth}{!}{%
\input{tabAverageResults}
%}
\end{center}
\end{table*}
\end{comment}

\begin{table*}[ht!]
%\centering
\footnotesize
\caption{Classification performance at different levels of the proposed semantic tree for food-related scenes categorization. We compute the achieved accuracy (Acc) per level and the weighted accuracy (W-Acc) where we consider the number of samples per class. The different semantic levels (L), Level 1 (L1), Level 2 (L2) and Level 3 (L3) are introduced in Fig. \ref{fig:baynet}.}
\label{table:tableLayersTree}
\input{tab_metaclasses}

\end{table*}

\begin{figure}[ht!]
\includegraphics[width=0.5\textwidth]{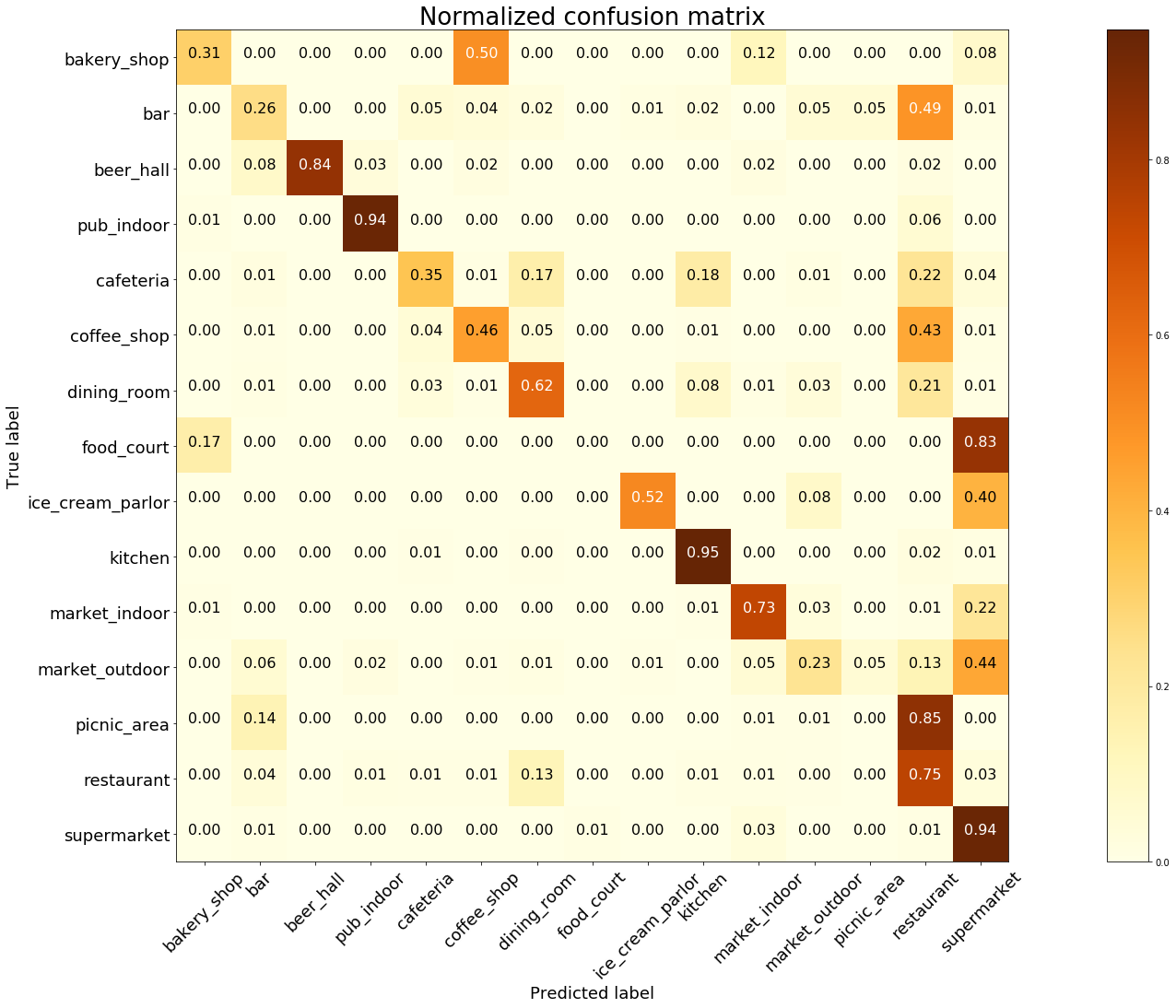}
\caption{Confusion matrix with the classification performance of the proposed hierarchical classification model.}
\label{fig:confusionmatrix}
\end{figure}

If we observe the confusion matrix in Fig. \ref{fig:confusionmatrix}, we can get insight about the miss-classified classes. We can see how our algorithm tends to confuse the classes belonging to the semantic level of \textit{self-service} (acquiring) and \textit{eating indoor} (eating). We believe that this is due to the unbalanced aspect of our data and the intrinsic similarity within the sub-categories of some of the branches of the semantic tree.

The classes with higher classification accuracy are \textit{kitchen} and \textit{supermarket}. We deduce that this is due to the very characteristic appearance of the environment that they involve and the number of different images of such classes in the dataset. On the contrary, \textit{picnic area} is not recognized by any of the methods. The confusion matrix indicates that the class is embedded by the model into the class \textit{restaurant}. This can be inferred by visually checking the images since in both classes a table and another person usually appear in front of the camera wearer. Moreover, from the obtained results, we can observe a relation between the previously computed Silhouette Score per class and the classification accuracy achieved by the classifiers. Classes with high consistency are better classified, while classes such as \textit{bar, bakery shop, picnic area}, or \textit{market outdoor} have lower classification performance.

\begin{figure*}
\centering
  \includegraphics[width=0.9\textwidth]{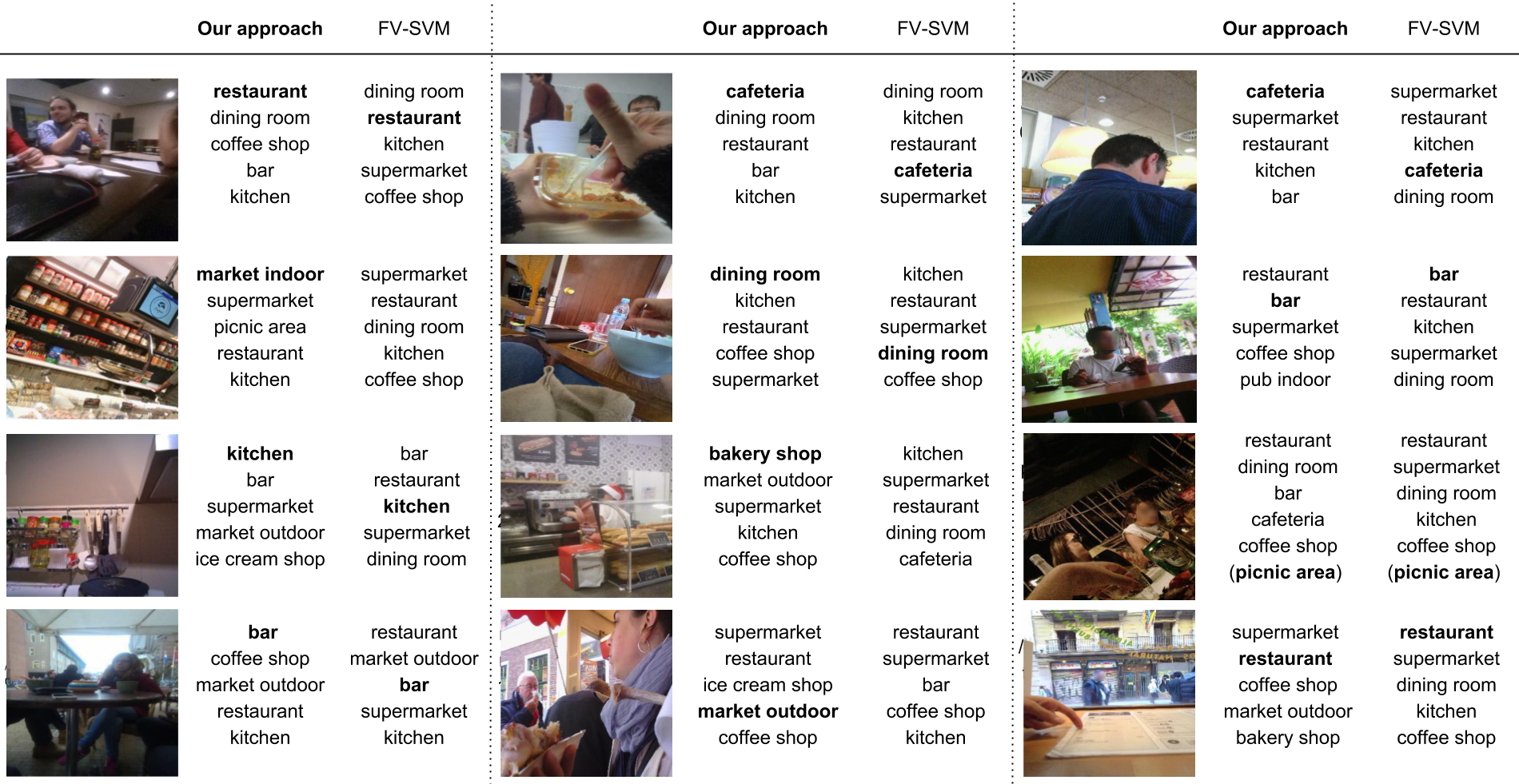}
  \caption{Examples of top 5 classes for the images in the test set. We show the results obtained by the proposed model, and compare them with the obtained ones by the trained SVM classifier. The class in bold corresponds to the true label of the given image.}
  \label{fig:qualitativeResults}
\end{figure*}

The achieved results are rather quantitatively similar. Therefore, we perform the \textit{t-test} to evaluate the statistical significance of the differences in performance. Our proposed model outperforms FV, SVMtree, FV+RF, FK+KNN, FV+SVM, MacNet, and ensembleCNNs with statistical significance ( p=$0.038*10^{-16}$, p=$0.042*10^{-12}$, p=$0.087*10^{-14}$, p=$0.057*10^{-13}$, p=$0.079*10^{-16}$, p=$0.087*10^{-19}$, and pvalue=$3.24*10^{-1}$ for paired t-test). The smaller the \textit{p} value, the higher the statistical significance.

From the results, we can discuss that the performance by the ensemble of CNNs is similar to the proposed model. This happens when it is evaluated at the level of image classification. We can see in Table II how the proposed hierarchy outperforms the baseline methods when classifying at the different levels of the taxonomy tree.

Qualitatively, in Fig. \ref{fig:qualitativeResults} we illustrate some correct and wrong classifications by our proposed model and the trained SVM (FV-SVM). We highlight the ground-truth class of the images in boldface. Even though the performance of the different tested models does not differ much, the proposed model has the ability to better generalize, as its weighted average accuracy indicates.

% SECTION %.....%.....%.....%.....%.....%.....%.....%.....%.....%.....%.....%.....%.....%.....%.....%.....
%.....%.....%.....%.....%.....%.....%.....%.....%.....%.....%.....%.....%.....%.....%.....%.....%.....

\section{Discussions}
% ---------- ---------- ---------- ---------- ---------- ---------- ----------
\label{section:Discussions}

%\subsection{Images vs. Events scene classification}
The proposed dataset is composed of manually selected images from recorded day photo-streams. These extracted images belong to food-related events, described as groups of sequential images representing the same scene. We find important to highlight that for the performed experiments, images belonging to the same event stayed together for either training or testing phase. Even though the classification of such scenes could have been events rather than images, we do not dispose of a higher number of events for the training phase in the case of event-based scene classification. The creation of a bigger egocentric dataset is a recurrent ongoing work. Next lines of work will address the analysis of events in order to study if they are connected and time-dependent.

%\subsection{Egocentric images for behaviour analysis}
Recorded egocentric images can be highly informative about the lifestyle, behaviour and habits of a person. In this work, we focus on the implementation of computer vision algorithms for data extraction from images. More specifically, on characterizing food-related scenes related to an individual for future assistance in controlling obesity and other eating disorders being of high importance for society. 
%However, other problems such as isolation of the wearer could be analyzed from these images. Performing food analysis from images captured when a person eats while using the wearable camera is highly challenging. 

Next steps could involve the analysis of other information e.g. the duration and regularity of nutritional activities. Based on extracted information regarding individuals, their daily habits can be extracted and characterized. The daily habits of people can be correlated to their personality since people's routine affects them differently. Moreover, within this context social relations and their relevance can be studied: the number of people a person sees per day, the length and frequency of their meetings and activities, etc and how social context influence people. All this information extracted from egocentric images is still to be studied in depth leading to powerful tools for an objective, long-term monitoring and characterization of the behaviour of people for better and longer life.

%\subsection{Applying BSN to other applications}
The introduced model can be easily extrapolated and implemented to other classification problems with semantically correlated classes. Organizing classes in a semantic hierarchy and embedding a classifier to each node of the hierarchy allow considering the estimated intermediate probabilities for the final classification.

The proposed model computes the final classification probability based on the aggregation of the probabilities of the different classification levels. The random probability of a given class is $1/|C|$, where $|C|$ is the number of children the parent class of that node has. Hence, having a high number of sub-classes (children nodes) for a specific node would tend to lower probability. There is a risk that a `wrong class node' gets higher final classification probability if it has few brother-sin the tree compared to the `correct class node'.

% ---------- ---------- ---------- ---------- ---------- ---------- ----------
% ---------- ---------- ---------- ---------- ---------- ---------- ----------
\subsubsection{Application to recorded days characterization}
% ---------- ---------- ---------- ---------- ---------- ---------- ----------
Food-related scenes recognition is very useful to get understanding of the patterns of behaviour of people. The presence of people at certain food-related places is of importance when describing their lifestyle and nutrition. While in this work we focus on the classification of such places, we use the labels given to the photo-streams to characterize the camera wearer's 'lived experiences' related to food. The characterization is given by the proposed model allows us to address the scene detection at different semantic levels. Thus, by using high-level information we increase the robustness and the level of the output information of the model.
 
\begin{figure}[ht!]
\includegraphics[width=1\linewidth]{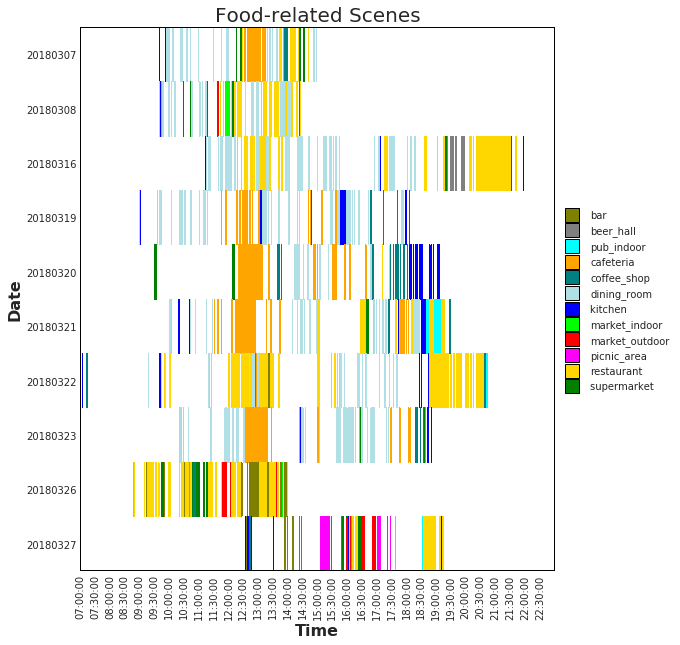}
\caption{Illustration of detected food-related events in egocentric photo-streams recorded during several days by the camera wearer.}
\label{fig:examplefinaloutput}
\end{figure}

In Fig. \ref{fig:examplefinaloutput}, we illustrate a realistic case where each row represents a recorded day by the camera wearer. As we have previously highlighted, our proposed model focuses on the classification of food-related scenes in egocentric photo-streams. However, the previous classification step would be the differentiating among Food and Non-food related images. In \cite{cartas2018batch} the authors addressed activity recognition in egocentric images. Thus, we apply their network and focus on images labelled as 'shopping' and 'eating or drinking', to later apply our proposed hierarchical model. In Fig. \ref{fig:examplefinaloutput} we can observe how not all labels are represented in the recorded days since it will depend on the life of the person. We can also monitor when the camera wearer goes for lunch to the \textit{cafeteria}, and conclude that s/he goes almost every day at the same time. We can recognize how \textit{restaurant} always occurs in the evening. With this visualization, we aim to show the consistency of the proposed tool for the monitoring of the time spent by the user at food-related scenes. The automatic and objective discovered information can be used for the improvement of the health of the user.

\vspace{-0.5cm}

% ---------- ---------- ---------- ---------- ---------- ---------- ----------
% ---------- ---------- ---------- ---------- ---------- ---------- ----------
\section{Conclusions}
% ---------- ---------- ---------- ---------- ---------- ---------- ----------
\label{section:conclusions}

In this paper, we introduced a multi-class hierarchical classification approach, for the classification of food-related scenes in egocentric photo-streams. The contributions of our presented work are three-fold: 
\begin{itemize}

\item A taxonomy of food-related environments that considers the main activities related to food (eating, cooking, buying, etc.). This semantic hierarchy aims to analyse the food-related activity at different levels of definition. This will allow a better understanding of the behaviour of the user. 

\item We propose a hierarchical model based on the combination of different layers of deep neural network, mirroring a given taxonomy for food-related scenes classification. This model is easily adapted to other classification problems and implemented on top of other different CNNs and traditional classifiers. The final classification of a given image is computed by combining the intermediate probabilities for the different levels of classification. Moreover, it showed its ability to classify images into meta-classes with high accuracy. This ensures that the final classification label, if not correct, will belong to a similar class.

\item A new dataset that we make publicly available. FoodEgoPlaces is composed of more than 33000 egocentric images describing 15 categories of food-related scenes of 11 camera wearers. We publish the data set as a benchmark in order to allow other scientists evaluating their algorithms and comparing their results with ours and with each other. We hope that future research addresses what we believe as a relevant topic: nutritional behaviour analysis in an automatic and objective way, by analysing the user’s daily habits from a first-person point of view.

\end{itemize}

The performance of the proposed architecture is compared with several built baseline methods. We use a pre-trained network on top of which we train our food-related scenes classifiers. However, transfer learning has shown its good performance when addressing problems where the lack of huge amounts of data is a problem. By building on top of pre-trained networks, we achieve results that outperform traditional techniques on the classification of egocentric images into challenging food-related scenes. Moreover and as an incentive, the proposed model has the ability of end-to-end automatically classifying different semantic levels of depth. Thus, specialists can analyze the nutritional habits of people and generate recommendations for improvement of their lifestyle by studying their food-related behaviour either from a broad perspective, such as when the person \textit{eats} or \textit{shops}, or into a more detailed one, like \textit{when  the person is eating in a fast-food restaurant}.

%\textcolor{orange}{The proposed model integrates a set of food-related activities, some of which can be analysed when addressing obesity disorders or cognitive impairment. On one side, the food environment influences obesity \cite{osei2012influence}. The analysis of the eating-routine of a person can help to control the diet better. For instance, someone could be interested in knowing the number of times per month eating somewhere specific (last layer of the taxonomy). For example, our system can help to quantify the time spent at fast-food restaurants, considered as poor-quality food environments, that have shown to negatively affect adolescents health \cite{jeffery2006fast}. On the other side, we can estimate the duration of custodian food-related activities (first layer of the taxonomy). Preparing meals or shopping as one of the main instrumental daily activities to evaluate cognitive decline \cite{morrow1999instrumental}.}

The analysis of the eating-routine of a person within its context/environment can help to control his/her diet better. For instance, someone could be interested in knowing the number of times per month that s/he goes to eat somewhere (last layer of the taxonomy). Moreover, our system can help to quantify the time spent at fast-food restaurants, that have shown to negatively affect adolescents health \cite{jeffery2006fast}. In a different clinical aspect, the capacities for preparing meal or shopping are considered as one of the main instrumental daily activities to evaluate cognitive decline \cite{morrow1999instrumental}. Our model allows analysing the custodian activities related to food-scenes represented in the first layer of the taxonomy. Hence, our proposed model integrates a set of food-related scenes and activities, that can boost numerous applications with very different clinical or social goals.

As future work, we plan to explore how to enrich our data using domain adaptation techniques. Domain adaptation allows the adaptation of the distribution of data to other target data distribution. Egocentric datasets tend to be relatively small due to the low-frequency rate of the recording cameras. We believe that by combining techniques of transfer learning, we will be able to explore how the collected dataset can be extrapolated to already available data, sets such as Places2. We expect that the combination of data distributions will improve the achieved classification performance. Therefore, further analysis of this line will allow us to get a better understanding of people's lifestyle, which will give insight into their health and daily habits.

% ---------- ---------- ---------- ---------- ---------- ---------- ----------
%\vspace{-0.5cm}
\section*{Acknowledgment}
% ---------- ---------- ---------- ---------- ---------- ---------- ----------

This work was partially funded by  projects TIN2015-66951-C2, SGR 1742, CERCA,  Nestore Horizon2020 SC1-PM-15-2017 (n° 769643), \textit{ICREA Academia 2014} and Grant 20141510 (Marat\'{o} TV3). The founders had no role in the study design, data collection, analysis, and preparation of the manuscript. The authors gratefully acknowledge the support of NVIDIA Corporation with the donation of several Titan Xp GPU used for this research. The collected data as part of the study and given labels is publicly available from the research group's website: \url{http://www.ub.edu/cvub/dataset/}

\bibliographystyle{IEEEtran}

\vspace{-1cm}

\begin{IEEEbiography}[{\includegraphics[width=1in,height=1.2in,clip,keepaspectratio]{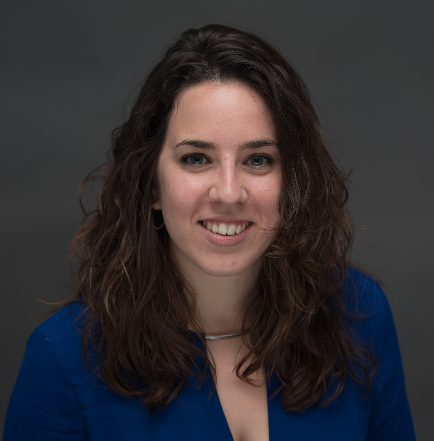}}]{Estefania Talavera}
received her BSc degree in electronic engineering from Balearic Islands University in 2012 and her MSc degree in biomedical engineering from Polytechnic University of Catalonia in 2014. She is currently a PhD student at the University of Barcelona and University of Groningen. Her research interests are lifelogging and health applications.
\end{IEEEbiography}

\vspace{-1.5cm}

\begin{IEEEbiography}[{\includegraphics[width=1in,height=1.2in,clip,keepaspectratio]{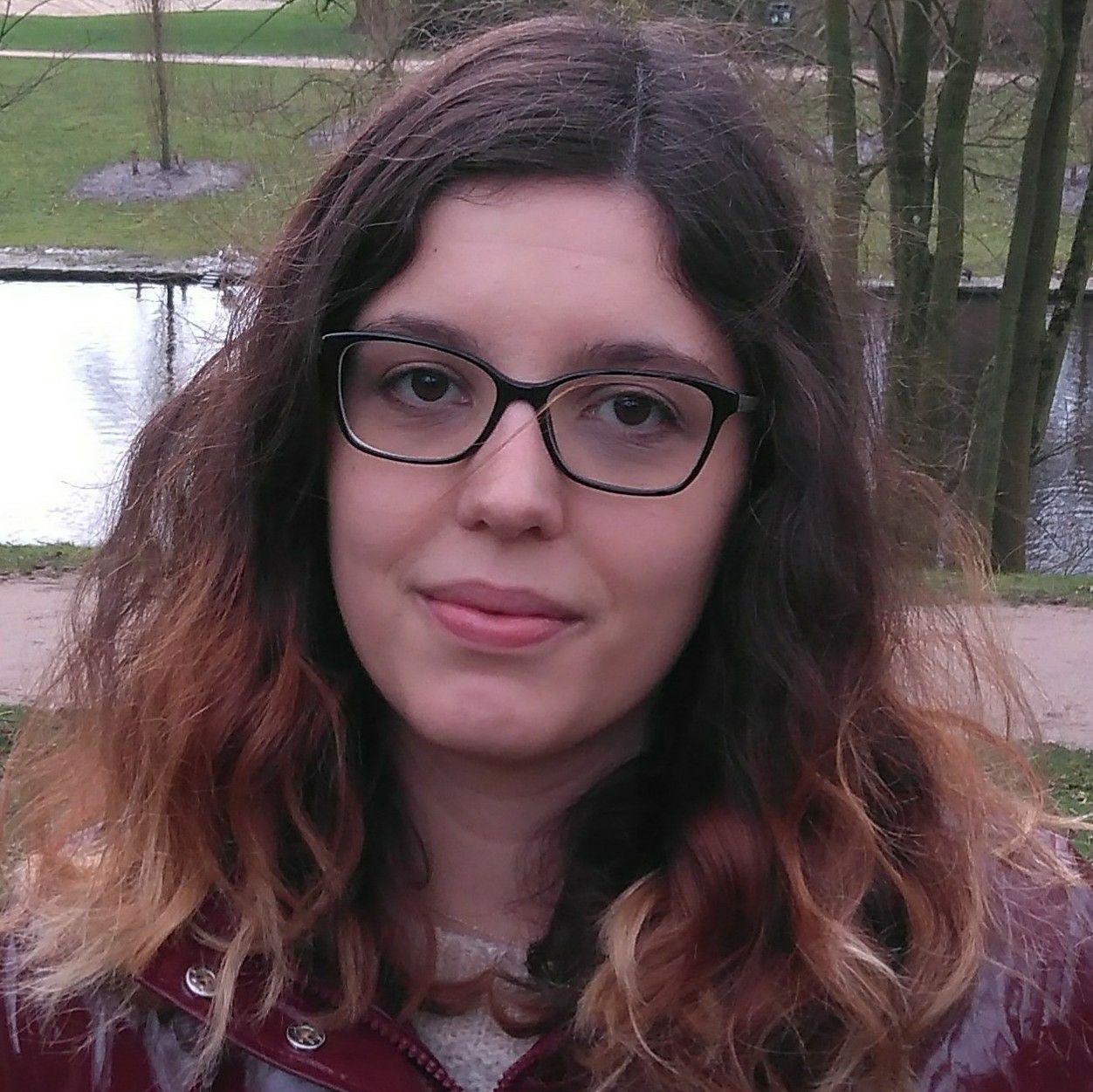}}]{María Leyva} received her BSc degree in Software Engineering from the University of Oviedo in 2011. She continued her studies with the MSc in Artificial Intelligence in the Polytechnic University of Catalonia, graduating in 2017. She is currently a PhD Student at the Johann Bernoulli Institute for Mathematics and Computer Science in the University of Groningen. \end{IEEEbiography}

\vspace{-1.5cm}

\begin{IEEEbiography}[{\includegraphics[width=1in,height=1.2in,clip,keepaspectratio]{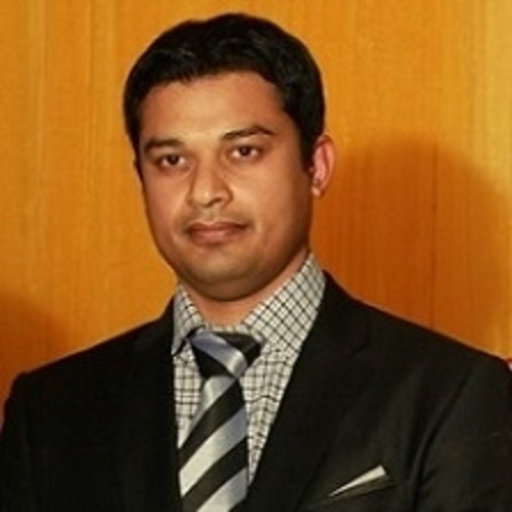}}]{Md. Mostafa Kamal Sarker} Md, Mostafa Kamal Sarker is a PhD student at the Intelligent Robotics and Computer Vision group, Rovira i Virgili Univerisity. He received his M.S. degree from Chonbuk National University, South Korea. His research interests include the areas of image processing, pattern recognition, computer vision, machine learning, deep learning, egocentric vision and visual lifelogging.
\end{IEEEbiography}

\vspace{-1.5cm}

\begin{IEEEbiography}[{\includegraphics[width=1in,height=1.2in,clip,keepaspectratio]{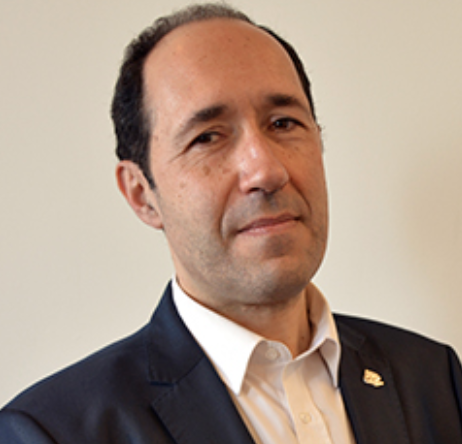}}]{Dr. Domenec Puig} 
received the M.S. and Ph.D. degrees in computer science from Polytechnic University of Catalonia, Barcelona, Spain. He is Associate Professor at the Department of Computer Science and Mathematics at Rovira i Virgili University, Spain. He is the Head of the Intelligent Robotics and Computer Vision group at the same university. His research interests include image processing, texture analysis, perceptual models for image analysis, scene analysis, and mobile robotics.
\end{IEEEbiography}

\vspace{-1.5cm}

\begin{IEEEbiography}[{\includegraphics[width=1in,height=1.2in,clip,keepaspectratio]{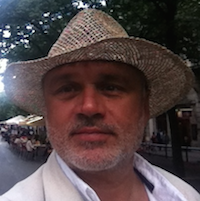}}]{Prof. Nicolai Petkov} 
received the Dr.Sc.Techn. degree in Computer Engineering from the Dresden University of Technology, Germany. He is Professor and Head of the Intelligent Systems group of the Bernoulli Institute for Mathematics, Computer Science and Artificial Intelligence of the University of Groningen, the Netherlands. His current research is in image processing, computer vision and pattern recognition, and includes computer simulations of the visual system of the brain, brain-inspired computing, and applications in health care and life sciences.
\end{IEEEbiography}

\vspace{-1.5cm}

\begin{IEEEbiography}[{\includegraphics[width=1in,height=1.2in,clip,keepaspectratio]{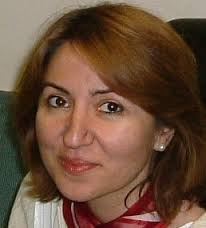}}]{Prof. Petia Radeva}
is a senior researcher and professor at the University of Barcelona (UB). She is Head of Computer Vision at the UB group and the MiLab of Computer Vision Center. Her present research interests are on the development of learning-based approaches for computer vision, egocentric vision and medical imaging.
\end{IEEEbiography}

\end{document}

%% file: figure_t-sne.tex
\begin{comment}
\minipage{0.4\textwidth}
\includegraphics[width=0.6\textwidth]{Figures/train_t-sne.png}
\endminipage\hfill
\minipage{0.4\textwidth}
\includegraphics[width=0.8\textwidth]{Figures/test_t-sne.png} 
\endminipage\hfill
\minipage{0.2\textwidth}%
\centering
\includegraphics[width=.7\textwidth]{Figures/legend_t-sne.png}
\endminipage
\end{comment}

%\includegraphics[width=.65\linewidth]{Figures/train_t-sne.png}
\includegraphics[width=\linewidth]{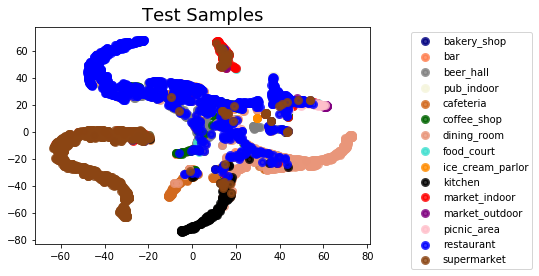}

%% file: table_acc_results.tex
\begin{tabular}{c|c|ccccccc}
& OurModel & FV & Tree+SVM & FV+RF & FV+SVM & FV+KNN & MACNet\cite{sarker2018macnet} & EnsembleCNNs \\  \hline \hline
bakery shop & 0.39 & 0.58 & 0.58 & 0.56 & 0.58 & 0.59 & 0.58 & 0.60 \\\hline
bar & 0.31 & 0.13 & 0.15 & 0.11 & 0.11 & 0.17 & 0.17 & 0.15 \\\hline
beer hall & 0.89 & 0.32 & 0.20 & 0.18 & 0.20 & 0.20 & 0.61 & 0.56 \\\hline
pub indoor & 0.85 & 0.70 & 0.71 & 0.71 & 0.71 & 0.71 & 0.64 & 0.82 \\\hline
cafeteria & 0.45 & 0.45 & 0.44 & 0.43 & 0.44 & 0.43 & 0.72 & 0.55 \\\hline
coffee shop & 0.59 & 0.40 & 0.39 & 0.34 & 0.38 & 0.34 & 0.49 & 0.49 \\\hline
dining room & 0.58 & 0.58 & 0.59 & 0.58 & 0.58 & 0.57 & 0.56 & 0.57 \\\hline
food court & 0.00 & 0.00 & 0.00 & 0.00 & 0.00 & 0.00 & 0.00 & 0.00 \\\hline
ice-cream parlor & 0.52 & 0.59 & 0.68 & 0.65 & 0.64 & 0.65 & 0.15 & 0.73 \\\hline
kitchen & 0.89 & 0.87 & 0.89 & 0.87 & 0.86 & 0.86 & 0.88 & 0.90 \\\hline
market indoor & 0.70 & 0.73 & 0.76 & 0.77 & 0.77 & 0.76 & 0.66 & 0.77 \\\hline
market outdoor & 0.28 & 0.20 & 0.20 & 0.20 & 0.20 & 0.19 & 0.23 & 0.25 \\\hline
picnic area & 0.00 & 0.00 & 0.00 & 0.00 & 0.00 & 0.00 & 0.00 & 0.00 \\\hline
restaurant & 0.70 & 0.67 & 0.68 & 0.68 & 0.68 & 0.68 & 0.63 & 0.73 \\\hline
supermarket & 0.85 & 0.81 & 0.81 & 0.79 & 0.81 & 0.80 & 0.75 & 0.84 \\  \hline \hline
Macro Precision & 0.56 & 0.53 & 0.55 & 0.59 & 0.55 & 0.56 & 0.48 & 0.60 \\\hline
Macro Recall & 0.53 & 0.48 & 0.47 & 0.44 & 0.46 & 0.45 & 0.49 & 0.52 \\\hline
Macro F1 & 0.53 & 0.47 & 0.47 & 0.46 & 0.46 & 0.46 & 0.47 & 0.53 \\  \hline \hline
Weighted Precision & 0.65 & 0.62 & 0.62 & 0.62 & 0.62 & 0.61 & 0.61 & 0.67 \\\hline
Weighted Recall & 0.68 & 0.64 & 0.64 & 0.64 & 0.64 & 0.64 & 0.63 & 0.68 \\\hline
Weighted F1 & 0.65 & 0.61 & 0.61 & 0.60 & 0.61 & 0.60 & 0.61 & 0.65 \\ \hline \hline
Accuracy & 0.68 & 0.64 & 0.64 & 0.64 & 0.64 & 0.64 & 0.63 & 0.68 \\\hline
Weighted Accuracy & 0.56 & 0.53 & 0.51 & 0.47 & 0.50 & 0.48 & 0.49 & 0.55 \\ \hline \hline
\end{tabular}

%% file: tab_metaclasses.tex
\resizebox{\textwidth}{!}{
\begin{tabular}{c|cc|cc|cc|cc|cc|cc|cc|cc|}
 & \multicolumn{2}{c|}{Our Method} & \multicolumn{2}{c|}{FV} & \multicolumn{2}{c|}{SVMTree} & \multicolumn{2}{c|}{FV+RF} & \multicolumn{2}{c|}{FV+SVM} & \multicolumn{2}{c|}{FV+KNN} & \multicolumn{2}{c|}{MACNet\cite{sarker2018macnet}} & \multicolumn{2}{c|}{EnsembleCNN} \\ \cline{2-17} 
 & Acc & WAcc & Acc & WAcc & Acc & WAcc & Acc & WAcc & Acc & WAcc & Acc & WAcc & Acc & WAcc& Acc & WAcc \\ \hline \hline
\multicolumn{1}{|c|}{Level 1 (L1)} & \textbf{0.944} & \textbf{0.947} & 0.927 & 0.919 & 0.934 & 0.931 & 0.928 & 0.922 & 0.927 & 0.924 & 0.927 & 0.910 & 0.884 & 0.865 & 0.923 & 0.913\\
\multicolumn{1}{|c|}{Level 2a (L2a)} & \textbf{0.915} & \textbf{0.685} & 0.886 & 0.664 & 0.898 & 0.673 & 0.890 & 0.666 & 0.800 & 0.753 & 0.890 & 0.648 & 0.829 & 0.623& 0.869 & 0.629 \\
\multicolumn{1}{|c|}{Level 2b (L2b)} & \textbf{0.893} & \textbf{0.947} & 0.890 & 0.940 & 0.890 & 0.944 & 0.885 & 0.945 & 0.897 & 0.935 & 0.885 & 0.927 & 0.860 & 0.906 & 0.856  & \textbf{0.955}\\ \hline \hline
\end{tabular}}